\documentclass[letterpaper, 10 pt, conference]{ieeeconf}  %

\usepackage[dvipsnames,table]{xcolor}

\usepackage[utf8]{inputenc}
\usepackage{graphicx}

\usepackage{amsmath}
\usepackage{subcaption}
\usepackage{multirow}
\usepackage{wrapfig}
\usepackage{xspace}
\usepackage{booktabs} 
\usepackage{amssymb}
\usepackage{braket}
\let\labelindent\relax
\usepackage{enumitem}
\usepackage{microtype}
\usepackage{arydshln}

\usepackage{adjustbox}
\usepackage{array}

\usepackage{pgfplots}
\pgfplotsset{compat=1.18}

\usepackage{hyperref}

\newif\ifshowedits

\newcommand{\addeditor}[3]{%
  \definecolor{#1color}{rgb}{#3}
  \expandafter\newcommand\csname #1\endcsname[1]{%
  \ifshowedits
    {\color{#1color} ##1}%
  \else
    {##1}%
  \fi
  }%
  \expandafter\newcommand\csname #1rmk\endcsname[1]{%
  \ifshowedits
    {\color{#1color} {\bf [#2: ##1]}}
  \fi
  }%
  \expandafter\newcommand\csname #1rpl\endcsname[2]{%
  \ifshowedits
    {\color{#1color} ##1 \sout{##2}}
  \else
    {##1}
  \fi
  }%
}

\newcommand{\ours}{TOAD\xspace}

\newcommand{\rev}[1]{{\color{black}#1}}

\addeditor{alex}{AB}{0.5, 0.5, 0.0}
\addeditor{andrei}{Andrei}{0, 0.5, 0.5}
\addeditor{ek}{EK}{.1,.7,.3}
\addeditor{eloi}{EZ}{0.2, 0.2, .75}
\addeditor{yuan}{YY}{0.6, 0.6, 0.0}
\addeditor{gilles}{GP}{0.0, 0.45, 0.60}
\addeditor{renaud}{RM}{1, 0.4, 0.0}

\addeditor{yihong}{YH}{0.2, 0.2, 1.0}

\showeditstrue

\newcolumntype{R}[2]{%
    >{\adjustbox{angle=#1,lap=\width-(#2)}\bgroup}%
    l%
    <{\egroup}%
}
\newcommand*\rot{\multicolumn{1}{R{45}{1em}}}%

\IEEEoverridecommandlockouts                              %

\title{\LARGE \bf
Test-Time Trajectory Optimization for Autonomous Driving
}

\author{
   Yihong Xu\textsuperscript{1,*} \quad
   Éloi Zablocki\textsuperscript{1,*} \quad
   Yuan Yin\textsuperscript{1} \quad
   Elias Ramzi\textsuperscript{1} \quad
   Ellington Kirby\textsuperscript{1} \quad \\
   {Alexandre Boulch}\textsuperscript{1} \quad
   {Matthieu Cord}\textsuperscript{1,2}
   \thanks{$^{1}$Valeo.ai, $^{2}$Sorbonne Université, CNRS, ISIR, F-75005 Paris, France }
   \thanks{corresponding author: yihong.xu@valeo.com}
   \thanks{*Equal contribution}
}

\begin{document}

\maketitle
\pagestyle{plain}

\begin{abstract}

End-to-end planners for autonomous driving typically generate a set of candidate trajectories, score each one, and return the highest-scoring candidate. However, the scorer is applied only after the proposals are generated and cannot influence the set of trajectories: a weak set of candidates limits planning performance regardless of the scorer's quality. We instead treat the scorer as a learned trajectory-level reward function and search for trajectories that maximize it. Our method, \ours, runs the Cross-Entropy Method at test time, warm-started from the planner's proposals. It requires no retraining and is plug-and-play for existing planners. Across seven base planners, \ours improves results on NAVSIM-v1 (94.7 PDMS), NAVSIM-v2 (56.3 EPDMS), and the closed-loop HUGSIM benchmark. The code is made publicly available via the project page: https://valeoai.github.io/TOAD/.

\end{abstract}

\section{Introduction}
\label{sec:intro}
End-to-end (E2E) planning has become the leading approach for autonomous driving \cite{bojarski2016pilotnet,hu2023uniad,renz2025simlingo}. 
A single network maps sensor data and ego state directly to a driving trajectory. Several state-of-the-art methods share a common recipe. They first generate a finite set of candidate trajectories, either drawn from a large pre-computed vocabulary \cite{li2024hydra,li2025gtrs,li2025ztrs,sun2026sparsedrivev2} or produced on the fly \cite{kirby2026drivor,feng2025rap,guo2025ipad}. A scorer then picks the best trajectory from this set. 

The scorer is at the core of the planning pipeline. It is trained to imitate an oracle of driving quality (safety, progress, comfort), and can therefore be seen as a learned, trajectory-level reward function, an objective that can be optimized. Yet current planners use it only once: generation is one-shot, and the scorer only ranks the fixed set of proposals. If every candidate is poor, the planner still returns the best of a bad set. We see this as a missed opportunity and ask whether the scorer can be used for more: not just \rev{to} rank a fixed set once, but \rev{to} guide a search for trajectories that score well.

We introduce \ours, a test-time procedure that turns a frozen scorer into the objective of a search over trajectories, at no training cost.
Concretely, \ours wraps a frozen E2E planner in an inference-time loop based on the Cross-Entropy Method (CEM) \cite{rubinstein2004cem}. At each iteration, we sample trajectories from a Gaussian distribution, score them with the frozen scorer, refit the Gaussian to the top-scoring candidates (`elites'), and repeat. The search is warm-started with the planner's selected trajectory and confined to its neighborhood, which limits how far the search can drift from the scorer's training distribution.

A key finding is that not all trained scorers can serve as rewards. Test-time search succeeds only when the scorer stays accurate beyond its training proposals, since it explores trajectories outside any fixed set. Scorers fit to a fixed vocabulary \cite{li2024hydra,li2025gtrs,li2025ztrs,sun2026sparsedrivev2} fail: they are accurate on their own candidates but degrade sharply elsewhere, and \rev{the search with them} drops below the baseline, despite looking strong under standard ranking evaluation. Only a disentangled scorer trained to evaluate freely decoded trajectories \cite{kirby2026drivor} works in \ours. Test-time search thus exposes a weakness in standard scorers that \rev{simple} ranking evaluation \rev{does not} reveal.

As it only requires a frozen planner and a generalizing scorer, \ours is plug-and-play. We attach a single frozen scorer to seven base planners, iPad \cite{guo2025ipad}, Hydra-MDP \cite{li2024hydra}, GTRS-A \cite{li2025gtrs}, ZTRS \cite{li2025ztrs}, RAP \cite{feng2025rap}, DiffusionDrive \cite{liao2025diffusiondrive}, and DrivoR \cite{kirby2026drivor}, bringing improvements on NAVSIM-v1 \cite{dauner2024navsim}, NAVSIM-v2 \cite{Cao2025navsim2}. We demonstrate that the gains come from the search itself: simply smoothing or re-ranking the base planner's proposals with the scorer can even hurt performance, whereas \ours improves it.
On NAVSIM-v2, DrivoR + \ours reaches 56.3 EPDMS, a new state of the art that nearly matches the privileged PDM-Closed (56.6) \cite{dauner2023pdm}. \ours also improves DrivoR's zero-shot \rev{driving score (HDS)} on the photorealistic closed-loop HUGSIM simulator \cite{zhou2026hugsim}.

Our contributions are as follows:
\begin{itemize}[noitemsep, topsep=0pt, leftmargin=1.2em]
    \item We reframe the E2E trajectory scorer as a learned reward function that can optimize trajectories. %
    \item We propose \ours, a test-time search that plugs into any base E2E planner with no retraining.
    \item We show that a successful optimization requires a scorer that generalizes beyond its training proposals, exposing a weakness in fixed-vocabulary scorers.
    \item Across seven planners, \ours gives consistent gains on NAVSIM-v1 and NAVSIM-v2, a new state of the art on NAVSIM-v2, and improved closed-loop performance on HUGSIM, at negligible cost.
\end{itemize}

\section{Related Work}

\noindent\textbf{End-to-End Driving and `Score-and-Select' Approaches.}
E2E learning is now dominant in autonomous driving. Early systems used complex pipelines of hand-designed modules for detection, tracking, and mapping \cite{zeng2019neural_motion_planner,yurtsever2020survey}. Recent methods instead map sensors and ego state to a trajectory with a single network \cite{hu2023uniad,jiang2023vad,paradrive}, often dropping 3D supervision and bird's-eye-view representations \cite{liao2025diffusiondrive,kirby2026drivor}. {Progress is measured on open-loop or pseudo closed-loop benchmarks such as NAVSIM-v1/v2 \cite{dauner2024navsim,Cao2025navsim2} and simulators like HUGSIM \cite{zhou2026hugsim}.}

The strongest E2E planners share a recipe: they generate candidate trajectories, score each, and output \rev{the highest-scoring one} \cite{li2024hydra}. Candidates come from a large vocabulary of pre-computed trajectories \cite{li2024hydra,li2025gtrs,li2025ztrs} or are generated on the fly, e.g., by diffusion or learnable queries \cite{liao2025diffusiondrive,xing2025goalflow,guo2025ipad,feng2025rap,kirby2026drivor}. Either way, the scorer is trained to imitate a driving-quality oracle and selects the trajectory by argmax. A clear trend is to enlarge the candidate set with ever denser vocabularies \cite{yao2025drivesuprim,sun2026sparsedrivev2}, which increases runtime latency \cite{kirby2026drivor} and often overfits the scorer to the fixed vocabulary. Further, oracle studies show an ideal selection can beat the human driver \cite{yao2025drivesuprim}, suggesting the candidate set, not the scorer, often limits performance. Some scorers are trained to transfer across vocabularies \cite{li2025gtrs}. However, they still rank a fixed set, which differs from staying accurate on freely searched trajectories. We instead ask whether a trained scorer can act as a reward function and search for trajectories beyond that set, further boosting performance.

\noindent\textbf{Test-Time Optimization and Search for Planning.}
\emph{Sampling-based optimization and MPC} have a long history in robotics and driving. Gradient-free samplers like the Cross-Entropy Method \cite{rubinstein2004cem,kobilarov2012cem,pinneri2021cem} and Model Predictive Path Integral control \cite{williams2017mppi} sample trajectories, score a cost, and refit toward low-cost regions, typically inside Model Predictive Control (MPC). Classical MPC for driving uses hand-designed costs; later work learns parts of it, e.g., cost maps from images \cite{drews2017aggressive} or cost weights in a differentiable planner \cite{huang2024dipp}. \rev{Closest to our setting}, UniAD \cite{hu2023uniad} post-optimizes its trajectory against a predicted occupancy map at inference. We share this search-at-test-time idea but optimize a fully learned end-to-end scorer, without predicting any intermediate representation such as future occupancy maps.

\emph{Test-time training} adapts the model itself rather than the trajectory. Centaur \cite{sima2025centaur} updates an E2E planner at test time by minimizing an uncertainty measure over its predictions. \rev{Our approach is} complementary and lighter: model weights stay frozen; we only search the trajectory space. %

\emph{Inference-time search} on a fixed model has driven large gains in language models, via inference-time reasoning \cite{openai2024O1,deepseek2025r1} and search guided by verifiers or reward models \cite{brown2024large}. This is appearing in driving too: re-ranking candidates with a vision-language model \cite{zheng2025svsf} or scoring them online with a learned world model \cite{li2025wote,sun2026rawmpc}. We also search at inference, but over a continuous trajectory space rather than a fixed set, using a single learned scorer rather than an auxiliary model.

\emph{Learned scorers as rewards} connect our setting to model-based RL, where a learned reward guides action selection. Several driving methods use a reward during training: RL fine-tuning of planners \cite{li2026recogdrive}, reinforced diffusion for generation \cite{song2025diver}, and RLHF for driving styles \cite{li2025learningpersonalizeddrivingstyles}. We instead use a frozen, already-trained scorer purely at inference, as the search objective, with no policy learning or change to the generator.

\begin{figure*}[ht]
    \centering
    \includegraphics[width=\linewidth]{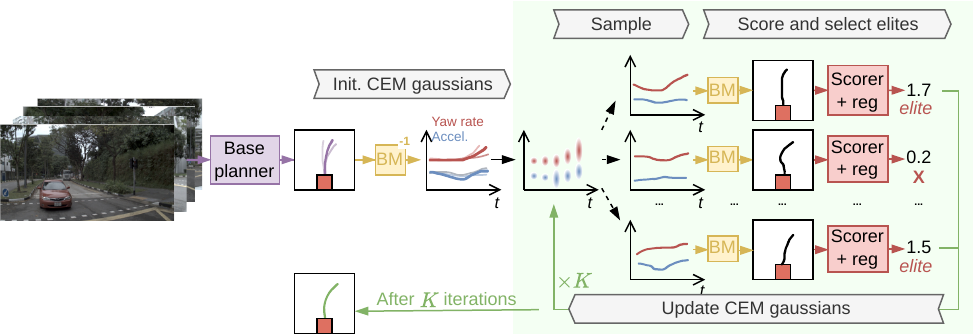}
    \caption{\textbf{Overview of \ours.} A frozen base planner outputs trajectory proposals and a selected anchor $\tau_{\text{base}}$ (left). 
    Trajectories are mapped to controls %
    by the inverse bicycle model ($\operatorname{BM}^{-1}$). 
    Then, a Cross-Entropy Method (CEM) loop runs for $K$ steps in control space: \rev{at each step it} \emph{samples} controls, rolls them out with the bicycle model ($\operatorname{BM}$), \emph{scores} them under `Scorer + reg' (the learned reward with comfort and anchor regularizers), \rev{and} refits the Gaussian to the highest-scoring \emph{elites}.} %
  \label{fig:method}
\end{figure*}
\section{Method}
\label{sec:method}
We present \ours, illustrated in \autoref{fig:method}, a scorer-based method that enhances trajectories produced by an arbitrary planner at test time. We introduce the scorer in \autoref{sec:reward} and then detail the optimization algorithm and its design choices in \autoref{sec:search}.

\subsection{Scorer as a Reward Function}
\label{sec:reward}

Our method uses a scorer $S$ as a reward function over trajectories. A trajectory $\tau = (p_1, \dots, p_{n_p})$ is a sequence of $n_p$ future poses $p_t = (x_t, y_t, \theta_t) \in \mathbb{R}^3$, \rev{with 2D coordinate $(x_t, y_t)$ and heading $\theta_t$ centered in the bird's-eye-view} ego frame. Conditioned on the observation $o$ (sensor inputs and ego state) and the goal $g$ (target or route), the scorer maps a complete trajectory to a scalar, $S(\tau; o, g) \in \mathbb{R}$. $S$ is a neural network trained jointly with a base planner. We treat $S$ as a \emph{reward} function over the trajectory space, which \rev{serves as} the action space. This enables searching for reward-maximal trajectories rather than ranking a fixed candidate set.

\subsection{Test-Time Trajectory Search}
\label{sec:search}

We instantiate this search with the Cross-Entropy Method (CEM)~\cite{rubinstein2004cem}, a classical sampling-based optimizer. To keep the search where the scorer is reliable, we confine it to a trust region around the base planner's proposals (\autoref{fig:method}, left). The following paragraphs describe and motivate each design choice; we ablate them individually in \autoref{sec:ablation}.

\noindent\textbf{Optimizing in control space.} For the sampling-based search, the choice of search space is not obvious. Sampling directly in pose space would yield jagged, dynamically infeasible trajectories, since it ignores vehicle dynamics entirely. We instead inject this prior through a kinematic bicycle model ($\operatorname{BM}$) and search over a sequence of controls,
\begin{equation}
u = (u_t)_{t=1}^{n_p}, \qquad u_t = (a_t, \omega_t) \in \mathbb{R}^2,
\end{equation}
where $a_t$ and $\omega_t$ are the per-step longitudinal acceleration and yaw rate. Our $\operatorname{BM}$ is invertible, so controls and trajectories are equivalent representations. From a control sequence $u$ we roll out a trajectory at the current ego speed $v_0$ as $\tau = \operatorname{BM}(u, v_0)$, and recover the controls by inverse kinematics, $u = \operatorname{BM}^{-1}(\tau, v_0)$, both illustrated in \autoref{fig:method}. This approach is standard in sampling-based control~\cite{williams2017mppi,kobilarov2012cem} and trajectory optimization~\cite{hanselmann2022king,yin2024regents}, and buys two properties for free: every sample is smooth and feasible, and comfort, defined on accelerations (the controls) and jerks (their derivatives), can be computed directly from $u$.

\noindent\textbf{Planner-based trust region.} In principle, CEM can run from any initialization (see \autoref{tab:cem_ablation} for ablations). We instead deliberately constrain the search to a small trust region around the base trajectory proposal conditioned on the observation $o$ and the navigation goal $g$, because the learned scorer of \autoref{sec:reward} is more reliable nearby. Outside this region, the search can drift toward trajectories that the scorer rates highly but cannot truly judge, which could lead to reward hacking.

Formally, we assume access to a trained \textit{base planner} that maps an observation $o$ and a goal $g$ to a set of $N$ candidate trajectories together with a selected trajectory,
\begin{equation}
\textsc{BasePlanner}(o, g) = (\mathcal{T}, \tau_{\text{base}}), \quad \tau_{\text{base}} \in \mathcal{T} = \{\tau^{(i)}\}_{i=1}^N.
\end{equation}
This formulation covers existing planners, including single-candidate ones where $\mathcal{T}=\{\tau_{\text{base}}\}$. Given the ego speed $v_0$, these map equivalently to control sequences, $\mathcal{U}=\{u^{(i)} = \operatorname{BM}^{-1}(\tau^{(i)}, v_0)\}_{i=1}^N$, with a selected one $u_{\text{base}}$. We use these to define the trust region, \rev{with $u_{\text{base}}$ as its anchor and the spread of $\mathcal{U}$ as its extent}.

\noindent\textbf{\ours: CEM with trust-region prior.}
In \ours, the trust-region prior shapes both the CEM \rev{objective} and its initialization. CEM maximizes $J(u)$, an objective combining $S$ and two closed-form regularizers:
\begin{equation}
J(u) = S\big(\operatorname{BM}(u, v_0); o, g\big) \;-\; \lambda_a\, C_{\text{anchor}}(u) \;-\; \lambda_c\, C_{\text{comf}}(u).
\end{equation}
The first regularizer, $C_{\text{anchor}}(u) = \lVert u - u_{\text{base}} \rVert^2$, ties the search to the trust region by penalizing deviation from the anchor.
The second, $C_{\text{comf}}$, accumulates squared violations of the standard kinematic comfort limits (longitudinal and lateral acceleration, jerk, yaw rate, and yaw acceleration); being exact and closed-form, it supplies a signal the learned scorer estimates only coarsely. We evaluate $C_{\text{comf}}$ over the future controls and the last executed control, penalizing uncomfortable transitions from the vehicle's recent motion.

To start the search, the CEM mean is set to the anchor, $\mu_0 = u_{\text{base}}$; the standard deviation starts from the spread of the base planner's own proposals, $\sigma_0 = \max(\beta \cdot \mathrm{std}(\mathcal{U}), \epsilon)$ with $\beta < 1$ and a small floor $\epsilon$ that prevents exploration from collapsing entirely. Exploration thus depends on the planner's own uncertainty.
We run CEM over controls with a Gaussian 
$\mathcal{N}(\mu_k, \mathrm{diag}(\sigma_k^2))$ in
$\mathbb{R}^{2 n_p}$. 
At iteration $k = 1, \dots, K$, we draw $M$ control sequences, roll them out, evaluate $J$, 
and keep the $E$ highest-scoring sequences as the elite set $\mathcal{E}_k$. We then refit the Gaussian to the elites:
\(
\mu_k = \mathrm{mean}(\mathcal{E}_k) \ \text{\rev{and}}\ \sigma_k = \mathrm{std}(\mathcal{E}_k).
\)

\begin{table*}[t]
\centering
\begin{minipage}[t]{0.68\textwidth}
   \centering
           \caption{\textbf{NAVSIM-v2 \texttt{navhard-two-stage}}. Comparison to existing methods on NAVSIM-v2~\cite{Cao2025navsim2}. \rev{Each sub-metric cell reads Stage~1\,/\,Stage~2. Higher is better.}
      }
    \scriptsize
      \setlength{\tabcolsep}{2.5pt}
            \resizebox{\columnwidth}{!}{
 \begin{tabular}{@{}l@{}r|ccccccccc|l@{}}
      \toprule
              & & \multicolumn{9}{c}{Stage 1 / Stage 2} & \\
      Method & & NC & DAC & DDC & TLC & EP & TTC & LK & HC & EC & \hspace{-0.25cm}EPDMS\\
      \midrule
  \rowcolor{black!10}
    PDM-Closed & \cite{dauner2023pdm} & 94.4/90.5 & 98.8/90.6 & 100/95.4 & 99.5/98.4 & 100/100 & 93.5/86.6 & 99.3/74.2 & 87.7/91.9 & 36.0/29.7 & 56.6 \\

    \midrule

    SpanVLA & \cite{zhou2026spanvla} & 98.4/86.9 & 94.3/84.3 & 97.8/87.1 & 99.9/98.2 & 85.7/85.5 & 97.2/82.7 & 94.2/62.3 & 97.6/96.8 & 72.1/67.4 & 40.1 \\
    DriveSuprim & \cite{yao2025drivesuprim} & 98.9/87.9 & 95.1/88.8 & 99.2/89.6 & 99.6/98.8 & 76.1/80.3 & 99.1/86.0 & 94.7/53.5 & 97.6/97.1 & 54.2/56.1 & 42.1 \\
    DiffVLA & \cite{jiang2025diffvla} & 95.7/81.2 & 99.2/88.8 & 100.0/94.6 & 100.0/99.0 & 85.9/86.0 & 96.4/76.4 & 97.1/59.8 & 95.0/98.6 & 84.2/80.4 & 45.0 \\
    GTRS-E & \cite{li2025gtrs} & 98.9/92.3 & 99.3/93.3 & 99.8/94.6 & 99.8/99.2 & 75.2/73.1 & 98.4/91.2 & 96.0/53.9 & 97.6/96.7 & 51.6/56.8 & 49.4 \\
    SimScale & \cite{tian2025simscale} & 99.6/94.5 & 99.1/94.2 & 99.9/95.8 & 100.0/99.2 & 69.6/75.8 & 99.6/92.8 & 95.8/60.1 & 95.6/96.1 & 28.4/43.2 & 53.2 \\
    DriveFuture & \cite{hong2026drivefuture} & 99.8/90.6 & 99.8/87.5 & 100.0/94.1 & 99.6/99.1 & 85.7/84.6 & 99.8/88.8 & 98.7/58.3 & 97.6/93.5 & 66.2/45.6 & 55.5 \\

    \midrule
        
    iPad & ~\cite{guo2025ipad} & 96.2/83.6 & 94.9/83.2 & 96.8/84.8 & 99.6/97.7 & 82.0/85.4 & 95.8/79.7 & 94.2/46.8 & 97.8/95.2 & 42.2/38.9 & 34.7 \\
      \rowcolor{blue!10}
    \textbf{\hspace{0.2cm}+ \ours} &  & \textcolor{blue}{97.7}/\textcolor{blue}{93.0} & \textcolor{blue}{97.1}/\textcolor{blue}{91.0} & \textcolor{blue}{99.6}/\textcolor{blue}{93.1} & \textcolor{blue}{99.8}/\textcolor{blue}{98.8} & 75.6/78.4 & \textcolor{blue}{98.4}/\textcolor{blue}{91.2} & 93.6/\textcolor{blue}{54.8} & 97.6/\textcolor{blue}{97.6} & \textcolor{blue}{43.1}/\textcolor{blue}{41.1} & \textcolor{blue}{49.8} \\

    RAP-DINO  & \cite{feng2025rap} & 97.1/83.2 & 94.4/83.9 & 98.8/87.4 & 99.8/98.0 & 83.9/86.9 & 96.9/80.4 & 94.7/52.3 & 96.4/95.2 & 66.2/52.4 & 39.6 \\
      \rowcolor{blue!10}
    \textbf{\hspace{0.2cm}+ \ours} &  & 97.1/\textcolor{blue}{89.9} & \textcolor{blue}{95.3}/\textcolor{blue}{89.7} & \textcolor{blue}{99.4}/\textcolor{blue}{92.5} & \textcolor{blue}{100.0}/\textcolor{blue}{98.2} & 80.3/81.4 & \textcolor{blue}{97.8}/\textcolor{blue}{87.7} & 92.9/52.1 & \textcolor{blue}{97.6}/\textcolor{blue}{97.2} & 64.4/51.1 & \textcolor{blue}{49.0} \\

    Hydra-MDP & \cite{li2024hydra} & 98.4/85.2 & 94.2/82.5 & 99.6/89.7 & 99.8/98.5 & 82.7/85.2 & 98.0/82.5 & 95.8/50.7 & 97.8/95.4 & 66.2/47.7 & 40.9 \\

    \rowcolor{blue!10}
    \textbf{\hspace{0.2cm}+ \ours} &  & 98.1/\textcolor{blue}{91.4} & \textcolor{blue}{95.1}/\textcolor{blue}{90.1} & 99.3/\textcolor{blue}{94.3} & \textcolor{blue}{100.0}/98.5 & 78.0/80.1 & \textcolor{blue}{98.2}/\textcolor{blue}{88.9} & 89.1/\textcolor{blue}{55.7} & 97.8/\textcolor{blue}{97.6} & 55.1/45.2 & \textcolor{blue}{49.7} \\

    GTRS-A & \cite{li2025gtrs} & 98.9/88.1 & 95.1/88.8 & 99.1/89.3 & 99.6/98.9 & 76.2/80.8 & 99.1/85.9 & 94.9/53.7 & 97.6/96.8 & 54.2/56.9 & 45.4 \\

   \rowcolor{blue!10}
    \textbf{\hspace{0.2cm}+ \ours} &  & 98.7/\textcolor{blue}{91.2} & \textcolor{blue}{96.4}/\textcolor{blue}{91.5} & \textcolor{blue}{99.2}/\textcolor{blue}{95.1} & \textcolor{blue}{99.8}/98.5 & 74.6/75.7 & 98.7/\textcolor{blue}{89.2} & 89.1/\textcolor{blue}{57.8} & 97.6/\textcolor{blue}{98.5} & \textcolor{blue}{60.9}/54.3 & \textcolor{blue}{51.7} \\

    ZTRS & \cite{li2025ztrs} & 98.9/91.1 & 97.6/90.4 & 100/95.8 & 100/99.0 & 66.7/63.6 & 98.9/89.8 & 96.2/60.4 & 96.7/97.6 & 44.0/66.1 & 48.1 \\

  \rowcolor{blue!10}
    \textbf{\hspace{0.2cm}+ \ours} &  & 98.7/\textcolor{blue}{92.3} & 96.9/90.2 & 99.6/94.0 & 99.6/98.5 & \textcolor{blue}{71.0}/\textcolor{blue}{67.8} & 98.4/\textcolor{blue}{90.4} & 90.7/58.8 & \textcolor{blue}{97.6}/\textcolor{blue}{99.2} & \textcolor{blue}{62.2}/57.3 & \textcolor{blue}{49.2} \\

    DrivoR & \cite{kirby2026drivor} & 99.1/92.3 & 98.2/91.6 & 99.3/97.3 & 99.8/99.1 & 75.4/75.7 & 98.7/90.6 & 94.9/56.1 & 97.6/98.4 & 70.2/44.7 & 54.6 \\
      \rowcolor{blue!10}
    \textbf{\hspace{0.2cm}+ \ours} &  & 98.9/\textcolor{blue}{93.6} & 98.2/\textcolor{blue}{94.0} & \textcolor{blue}{99.4}/\textcolor{blue}{97.5} & \textcolor{blue}{100.0}/98.8 & 74.6/74.1 & \textcolor{blue}{98.9}/\textcolor{blue}{92.0} & 94.0/\textcolor{blue}{57.1} & 97.6/98.0 & 66.7/\textcolor{blue}{47.3} & \bf \textcolor{blue}{56.3} \\
      \bottomrule

      \end{tabular}
            }
      \label{tab:benchmark_navsim_v2}
\end{minipage}
\hfill
\begin{minipage}[t]{0.3\textwidth}
        \scriptsize
      \centering
          \caption{\textbf{NAVSIM-v1} comparisons to different methods (\texttt{navtest} ($\uparrow$)).
      }
      \setlength{\tabcolsep}{1.8pt}
      \resizebox{\columnwidth}{!}{
      \begin{tabular}{@{}l@{}r|ccccc|l@{}}
      \toprule
      Method & & NC & DAC & TTC & Comf. & EP & \hspace{-0.1cm}{PDMS} \\
      \midrule
      \rowcolor{black!10}
      PDM-Closed & \cite{dauner2023pdm}               & 94.6 & 99.8 & 89.9 & 86.9 & 99.9 & 89.1 \\
      \rowcolor{black!10}
      Human driver& \cite{dauner2024navsim}           & 100 & 100&  100 & 99.9 & 87.5 & 94.8 \\

      \midrule

      DriveVLA-W0 & \cite{li2025drivevla}             & 98.7 & 99.1 & 95.3 & 99.3 & 83.3 & 90.2 \\
      ReCogDrive & \cite{li2026recogdrive}            & 97.9 & 97.3 & 94.9 & 100  & 87.3 & 90.8 \\
      Hydra-MDP++   & \cite{li2025hydramdppp}         & 98.6 & 98.6 & 95.1 & 100  & 85.7 & 91.0 \\

      Centaur & \cite{sima2025centaur}                & 99.5 & 98.9 & 98.0 & 100  & 85.9 & 92.6 \\

      DriveSuprim  & \cite{yao2025drivesuprim}        & 98.6 & 98.6 & 95.5 & 100 & 91.3 & 93.5 \\
      
    \midrule

    ZTRS & \cite{li2025ztrs}           & 98.8 & 99.5 & 95.1 & 100 & 74.7 & 86.9      \\
  \rowcolor{blue!10}
  \textbf{\hspace{0.2cm}+ \ours} &                                & 98.4 & 98.6 & 94.3 & 100 & \textcolor{blue}{82.1} & \textcolor{blue}{89.0}      \\

    DiffusionDrive & \cite{liao2025diffusiondrive}    &  98.2 & 96.2 & 94.7 & 100 & 82.2 & 88.1       \\
  \rowcolor{blue!10}
  \textbf{\hspace{0.2cm}+ \ours} &                                & 99.0 & 98.7 & 96.4 & 100 & 87.0 & \textcolor{blue}{92.2}      \\

    GTRS-A & \cite{li2025gtrs}    & 99.0 & 98.9 & 95.9 & 100 & 82.8 & 90.4      \\
       \rowcolor{blue!10}
       \textbf{\hspace{0.2cm}+ \ours} &                                & 98.9 & 98.1 & \textcolor{blue}{96.3} & 100 & \textcolor{blue}{84.8} & \textcolor{blue}{90.9}      \\

    Hydra-MDP & \cite{li2024hydra}                    & 98.5 & 98.7 & 94.9 & 100 & 85.5 & 90.9      \\
  \rowcolor{blue!10}
  \textbf{\hspace{0.2cm}+ \ours} &                                & \textcolor{blue}{98.7} & 98.0 & \textcolor{blue}{95.4} & 100 & \textcolor{blue}{87.2} & \textcolor{blue}{91.7}      \\

    iPad & \cite{guo2025ipad}                         & 98.6 & 98.3 & 94.9 & 100  & 88.0 & 91.7 \\
       \rowcolor{blue!10}
       \textbf{\hspace{0.2cm}+ \ours} &                                & \textcolor{blue}{99.0} & \textcolor{blue}{99.3} & \textcolor{blue}{96.4} & 100 & \textcolor{blue}{89.1} & \textcolor{blue}{93.4}      \\

    RAP-DINO & \cite{feng2025rap}                     & 99.1 & 98.9 & 96.7 & 100 & 90.3 & 93.8 \\
       \rowcolor{blue!10}
       \textbf{\hspace{0.2cm}+ \ours} &                                & 99.0 & \textcolor{blue}{99.2} & 96.6 & 100 & 90.1 & \textcolor{blue}{94.1 }      \\

    DrivoR & \cite{kirby2026drivor}             & 99.1 & 99.2 & 96.9 & 100 & 91.6 & 94.6      \\
       \rowcolor{blue!10}
       \textbf{\hspace{0.2cm}+ \ours}  &                               & 99.0 & \textcolor{blue}{99.3} & 96.8 & 100 & \textcolor{blue}{91.8} & \textcolor{blue}{\bf 94.7}      \\
      \bottomrule
      \end{tabular}
      }
      \label{tab:benchmark_navsim_v1}
  \end{minipage}

\end{table*}

After the final iteration, \rev{following standard CEM practice,} we roll out the trajectory induced by the converged
mean, $\tau_{\text{mean}} = \operatorname{BM}(\mu_K, v_0)$. We return whichever of \rev{$\tau_{\text{mean}}$} and the base proposal $\tau_{\text{base}}$ maximizes the \rev{objective},
\begin{equation}
\tau^\star = \operatorname*{arg\,max}_{\tau \in \{\tau_{\text{mean}},\, \tau_{\text{base}}\}}
\; S(\tau; o, g) - \lambda_c\, C_{\text{comf}}\bigl(\operatorname{BM}^{-1}(\tau, v_0)\bigr).
\end{equation}
$C_{\text{anchor}}$ is excluded here: the CEM mean should not be penalized for deviating from the anchor.
Comparing against $\tau_{\text{base}}$ guarantees the returned trajectory never \rev{scores lower} than $\tau_{\text{base}}$. 

\noindent\textbf{Scorer choice.} CEM requires a scorer that stays accurate off the base planner's proposal distribution, which is why \ours uses DrivoR's disentangled scorer~\cite{kirby2026drivor}. We highlight one property of our setup: the \emph{same} scorer serves every base planner and need not be the planner's own.

\section{Experimental Results}
\label{sec:experiments}

We evaluate our test-time search on standard driving benchmarks. 
{\ours consistently improves various base planners on \rev{the} real-world logs of NAVSIM-v1/2, \rev{whose scores correlate strongly with closed-loop evaluation}~\cite{Cao2025navsim2}, setting a new state of the art on both benchmarks.
It further improves over DrivoR on the closed-loop HUGSIM\rev{, and we present qualitative and failure analyses} (\autoref{sec:main-results}).}
\rev{We study the choice of scorer for} \ours in \autoref{sec:source}. \rev{Computation cost is studied in~\autoref{sec:footprint}}. Finally, \rev{we ablate the design choices} (\autoref{sec:ablation}).

\subsection{Experimental Setup}
\label{sec:setup}
\noindent\textbf{Benchmarks and metrics.} 
We evaluate on NAVSIM-v1~\cite{dauner2024navsim} (\texttt{navtest} split) and NAVSIM-v2~\cite{Cao2025navsim2} (\texttt{navhard-two-stage} split), two pseudo-closed-loop benchmarks built on real driving logs\rev{, scored by the official PDMS and EPDMS. PDMS aggregates no-at-fault collisions (NC), drivable area compliance (DAC), time to collision (TTC), ego progress (EP), and comfort (Comf.); EPDMS is computed over two evaluation stages and adds driving direction compliance (DDC), traffic light compliance (TLC), and lane keeping (LK), with history comfort (HC) and extended comfort (EC) in place of comfort}.
We also evaluate on closed-loop HUGSIM~\cite{zhou2026hugsim}, with photorealistic scenes constructed using 3D Gaussian Splatting~\cite{kerbl3Dgaussians}\rev{, scored by the HUGSIM Driving Score (HDS) over NC, DAC, TTC, comfort (Comf.), and route completion (RC)}. 
For the ablation, we use \texttt{warmup-two-stage}\footnote{\texttt{warmup-two-stage} intersects with \texttt{navhard-two-stage}\rev{; upon request, the benchmark authors validated its use as a validation set}.} as \rev{a validation set}.\\[0.5em]
\noindent\textbf{Base planners.} 
We show that \ours is {\textit{plug-and-play}} by evaluating it on seven base planners spanning \rev{three} proposal styles, i.e., fixed-vocabulary scorers (Hydra-MDP~\cite{li2024hydra}, GTRS-A~\cite{li2025gtrs}, ZTRS~\cite{li2025ztrs}), \rev{a diffusion-based planner (DiffusionDrive~\cite{liao2025diffusiondrive})\footnote{DiffusionDrive needs LiDAR, missing in NAVSIM-v2.},} and on-the-fly generators (iPad~\cite{guo2025ipad}, RAP~\cite{feng2025rap}, DrivoR~\cite{kirby2026drivor}). For each, we use the public checkpoint and its released trajectory proposals; no models are updated. \rev{For the ablation study, we choose iPad for its simplicity and anchor-/scorer-independence.} \\[0.5em]
\noindent\textbf{Hyperparameters.}
Unless stated otherwise, we run $K = 5$ CEM iterations for planners with on-the-fly/diffusion-based generators and $K = 100$ for those with fixed-vocabulary scorers due to \rev{their} much larger vocabulary ($>$8K) that initializes the search. The per-iteration candidate count $M$ is fixed to 64 \rev{to bound the search cost}, and the elite count to $E = M / 8$. We use an initial-std scale of $\beta = 0.5$, floored at $\epsilon_a = 0.1~\mathrm{m/s^2}$ and $\epsilon_\omega = 0.025~\mathrm{rad/s}$. The objective uses comfort weight $\lambda_c = 0.05$ and anchor weight $\lambda_a = 0.5$. All values above are kept fixed across base planners and benchmarks\rev{, and are ablated} in \autoref{sec:ablation}. \rev{\ours is run with a fixed seed (2), making it deterministic. Result variance and significance are studied in~\autoref{sec:main-results}.}

\subsection{Main results}
\label{sec:main-results}
\noindent\textbf{NAVSIM.} \ours improves every base planner on \rev{both} NAVSIM benchmarks. On NAVSIM-v2 (\autoref{tab:benchmark_navsim_v2}) the gains range from +2.3\% (ZTRS) to +43.5\% (iPad), and they compress the ranking: 
the six planners \rev{evaluated on NAVSIM-v2} start almost 20 EPDMS points apart (34.7 to 54.6) but all land between 49.0 and 56.3 after search. 
The weakest planners gain the most, while DrivoR, already the strongest, gains +3.1\% to reach 56.3 EPDMS\rev{. It outperforms} the strongest learned method (DriveFuture \cite{hong2026drivefuture}, 55.5) %
and is only 0.3 behind the privileged PDM-Closed (56.6), which uses ground-truth perception. %
The search \rev{targets} whichever sub-score has room to improve. On the harder, out-of-distribution NAVSIM-v2 scenes it trades a little progress for large safety gains in no-collision, drivable-area, and time-to-collision (\autoref{tab:benchmark_navsim_v2}). 
On the near-saturated NAVSIM-v1, where safety is already high, it instead extends progress \rev{(\autoref{tab:benchmark_navsim_v1})}. \rev{Despite} the saturated performance of DrivoR, \ours pushes it to 94.7 PDMS, the top entry \rev{overall}.
\rev{The search is deterministic only with a fixed seed (set to 2 in all experiments) and hyperparameters, we re-check that these gains are not seed noise.}
\autoref{tab:variance} re-runs four different planners (fixed-vocabulary and on-the-fly planners) over seeds 1 to 5: the standard deviation never exceeds 0.41 points, every mean $\pm$ std stays above the base planner, and a Wilcoxon signed-rank test~\cite{wilcoxon1945individual} on per-scenario deltas confirms each gain ($p \le 5{\times}10^{-5}$).

\begin{table}[b]
\centering
\caption{\textbf{Seed variance and significance.} Mean $\pm$ std of \ours{} over 5 seeds on NAVSIM-v1 (PDMS) and NAVSIM-v2 (EPDMS); $p$: Wilcoxon signed-rank test on per-scenario deltas between \ours{} and the base planner.}
\label{tab:variance}
\small
\setlength{\tabcolsep}{3pt}
\resizebox{\columnwidth}{!}{%
\begin{tabular}{lcccccc}
\toprule
& \multicolumn{3}{c}{NAVSIM-v1 (PDMS)} & \multicolumn{3}{c}{NAVSIM-v2 (EPDMS)} \\
\cmidrule(lr){2-4} \cmidrule(lr){5-7}
Planner & Base & +\ours & $p$ & Base & +\ours & $p$ \\
\midrule
Hydra-MDP~\cite{li2024hydra}    & 90.9 & 91.7 $\pm$ 0.15 & $<\!10^{-10}$ & 40.9 & 49.2 $\pm$ 0.41 & $5{\times}10^{-5}$ \\
iPad~\cite{guo2025ipad}         & 91.7 & 93.3 $\pm$ 0.03 & $<\!10^{-10}$ & 34.7 & 50.4 $\pm$ 0.41 & $<\!10^{-10}$ \\
DrivoR~\cite{kirby2026drivor}   & 94.6 & 94.8 $\pm$ 0.08 & $<\!10^{-10}$ & 54.6 & 56.2 $\pm$ 0.15 & $8{\times}10^{-10}$ \\
RAP-DINO~\cite{feng2025rap}     & 93.8 & 94.2 $\pm$ 0.05 & $<\!10^{-10}$ & 39.6 & 49.4 $\pm$ 0.20 & $<\!10^{-10}$ \\
\bottomrule
\end{tabular}%
}
\end{table}

\begin{table*}[t]
    \scriptsize
    \setlength{\tabcolsep}{1.5pt}
    \centering
    \caption{\textbf{HUGSIM~\cite{zhou2026hugsim}} zero-shot scores. \ours improves safety (NC, DAC, TTC) and comfort, raising \rev{HDS}, while route completion (RC) sometimes drops, indicating more cautious driving.}
    \label{tab:hugsim}
    \resizebox{\textwidth}{!}{
\begin{tabular}{@{}lr cccccc cccccc cccccc cccccc@{}}
\toprule
&& \multicolumn{6}{c}{KITTI360 \cite{kitti360}}
& \multicolumn{6}{c}{nuScenes \cite{nuscenes}}
& \multicolumn{6}{c}{Pandaset \cite{pandaset}}
& \multicolumn{6}{c}{Waymo \cite{waymo}} \\
\cmidrule(lr){3-8}\cmidrule(lr){9-14}\cmidrule(lr){15-20}\cmidrule(lr){21-26}
&&
NC & DAC & TTC & Comf. & RC & HDS &
NC & DAC & TTC & Comf. & RC & HDS &
NC & DAC & TTC & Comf. & RC & HDS &
NC & DAC & TTC & Comf. & RC & HDS \\
\midrule

LTF& \cite{chitta2022transfuser} & 
32.4 & 86.4 & 25.1 & 97.3 & 19.4 & 8.0 & 
56.1 & 99.7 & 49.8 & 95.4 & 50.1 & 35.1 & 
44.7 & 100.0 & 36.2 & 92.1 & 51.3 & 28.2 & 
49.1 & 93.9 & 42.1 & 95.4 & 38.3 & 23.8 \\

UniAD &\cite{hu2023uniad} & 
48.1 & 85.1 & 19.8 & 37.2 & 23.5 & 5.9 & 
71.8 & 99.2 & 63.3 & 79.3 & 49.1 & 37.5 & 
68.2 & 100.0 & 58.1 & 77.3 & 49.5 & 36.7 & 
76.9 & 94.3 & 66.3 & 71.3 & 56.2 & 44.4 \\

VAD &\cite{jiang2023vad} & 
24.4 & 83.2 & 15.0 & 94.2 & 17.1 & 3.9 & 
57.2 & 97.6 & 42.9 & 97.4 & 43.1 & 24.5 & 
42.4 & 100.0 & 26.5 & 94.9 & 38.7 & 15.4 & 
46.6 & 85.5 & 32.6 & 94.9 & 29.4 & 11.0 \\ 

GTRS-D & \cite{li2025gtrs} &
48.2 & 90.8 & 43.7 & 94.1 & 28.3 & 18.9 &
70.4 & 99.8 & 63.7 & 96.3 & 58.5 & 46.7 &
56.0 & 100.0 & 48.8 & 91.9 & 54.2 & 36.6 &
65.3 & 96.8 & 60.9 & 95.0 & 52.3 & 40.9 \\

GTRS-A & \cite{li2025gtrs} &
39.8 & 86.1 & 33.8 & 95.5 & 27.5 & 15.3 &
61.0 & 99.9 & 54.7 & 94.6 & 58.4 & 42.9 &
51.6 & 100.0 & 43.3 & 95.2 & 53.0 & 33.4 &
58.8 & 96.7 & 52.5 & 95.7 & 47.1 & 33.4 \\

DrivoR & \cite{kirby2026drivor} & 
44.0 & 89.9 & 40.8 & 93.6 & 28.8 & 17.7 & 
57.0 & 100.0 & 51.3 & 93.0 & 54.2 & 39.7 & 
50.8 & 99.9 & 44.0 & 89.8 & 58.3 & 35.1 & 
64.0 & 96.8 & 58.6 & 94.1 & 56.5 & 44.2 \\

\rowcolor{blue!10}\multicolumn{2}{l}{DrivoR \textbf{+ \ours}}  & \textcolor{blue}{49.5} & \textcolor{blue}{90.9} & \textcolor{blue}{46.1} & \textcolor{blue}{99.1} & 26.9 & \textcolor{blue}{18.4} & \textcolor{blue}{69.1} & 100.0 & \textcolor{blue}{63.9} & \textcolor{blue}{98.3} & \textcolor{blue}{60.9} & \textcolor{blue}{49.9} & \textcolor{blue}{57.3} & \textcolor{blue}{100.0} & \textcolor{blue}{51.6} & \textcolor{blue}{98.4} & 54.5 & \textcolor{blue}{38.0} & \textcolor{blue}{70.3} & 96.7 & \textcolor{blue}{66.6} & \textcolor{blue}{97.9} & 51.1 & 42.8 \\

\bottomrule
    \end{tabular}
    }
\end{table*}

\begin{figure*}[t]
    \centering
    \begin{subfigure}{0.49\linewidth}
        \centering
\includegraphics[width=\textwidth, trim=100pt 0 0 0, clip]{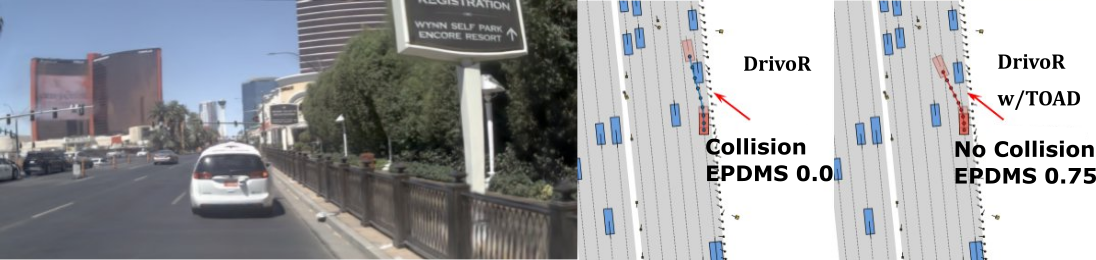}
        \caption{Success case}
        \label{fig:success}
    \end{subfigure}\hfill
    \begin{subfigure}{0.49\linewidth}
        \centering
\includegraphics[width=\textwidth, trim=100pt 0 0 0, clip]{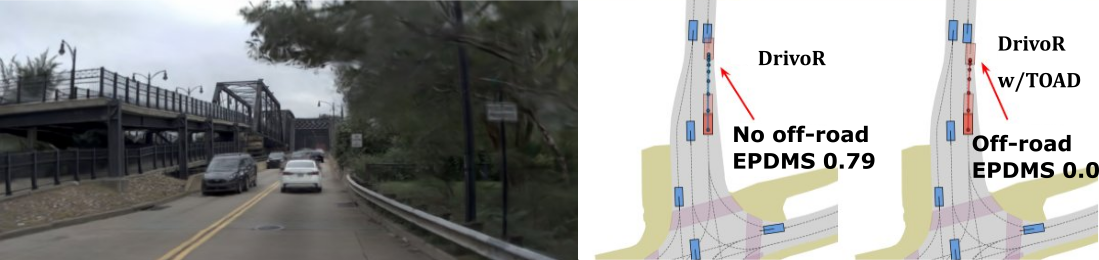}
        \caption{Failure case}
        \label{fig:failure}
    \end{subfigure}
    \caption{\textbf{\rev{Example improvement and failure visualization}} found w/ and w/o \ours on top of DrivoR~\cite{kirby2026drivor}.}
    \label{fig:success_failure_comparison}
\end{figure*}

\smallskip\noindent\textbf{HUGSIM \cite{zhou2026hugsim}.}
\autoref{tab:hugsim} reports zero-shot closed-loop evaluation, applying \ours on top of DrivoR. The safety sub-metrics (no-collision, drivable-area, time-to-collision) and, above all, comfort improve on nearly all four datasets, the comfort gain \rev{being} a direct effect of the control-space search. These gains raise the HUGSIM Driving Score (HDS) where there is room, most clearly on nuScenes (DrivoR: 39.7 to 49.9) and Pandaset (35.1 to 38.0). Route completion tends to drop, showing that the searched policy drives more cautiously, and on Waymo this caution leaves the score slightly lower. That the improvements transfer to a closed-loop simulator with no tuning suggests they reflect genuinely safer driving.

\smallskip\noindent\textbf{Qualitative examples and failure analysis.} We show in~\autoref{fig:success} \rev{that the refined trajectory avoids a collision with the car in front. \ours produces a better trajectory that is absent from the original proposal set.} 
\rev{As \autoref{fig:mpc_ablation_iterations} also suggests,} the reward is not always precise and can occasionally fail. Such failures may occur when visual cues are unreliable due to limited visibility, \rev{as} illustrated in \autoref{fig:failure}. \rev{Quantitatively, \ours falls below the base planner on 18--28\% (NAVSIM-v1) / 21--35\% (NAVSIM-v2) of scenarios across DrivoR, Hydra-MDP, and iPad, and severe drops ($\Delta < -0.1$) affect only 2--4\% / 5--17\%. Most regressions are shallow progress dips with no safety metric touched. The severe cases fall into three modes: (1) DAC: lateral maneuvers near hard geometry (narrow corridors), dominant on NAVSIM-v2; (2) TTC/NC: close following or crossing traffic, dominant on NAVSIM-v1 and in blurry synthetic scenes on NAVSIM-v2; (3) EP: dense slow interactive scenes, where \ours cedes progress.}

\begin{table}[t]
 \setlength{\tabcolsep}{1.8pt}
\centering
\caption{\textbf{What drives the gain?} \rev{EPDMS on NAVSIM-v2. \ours{} vs.\ (i) smoothing via the bicycle model (BM), (ii) re-scoring, (iii) searching w/ different scorers.}}
  \label{tab:w_wo_MPC_small}
\scriptsize
\resizebox{0.7\linewidth}{!}{%
  \begin{tabular}{@{}l l@{}}
    \toprule
    Method & EPDMS $(\uparrow)$ \\
    \midrule
    iPad~\cite{guo2025ipad} & 34.7 \\
    + Smoothing (BM)              &{35.0 ($\uparrow$\phantom{0}0.9\%)} \\
    + Re-scoring w/ DrivoR        & {45.6 ($\uparrow$31.4\%)} \\
    + Re-scoring w/ GTRS-A          & 34.5 ($\downarrow$\phantom{0}0.6\%) \\
    + Re-scoring w/ SparseDriveV2 & 30.3 ($\downarrow$12.7\%) \\
    + Search w/ GTRS-A scorer       & 23.9 ($\downarrow$31.1\%) \\
    + Search w/ SparseDriveV2 scorer & 34.6 ($\downarrow$\phantom{0}0.3\%) \\
    \rowcolor{blue!10}
    \textbf{+ \ours{} (search w/ DrivoR scorer)} & \textcolor{blue}{49.8 ($\uparrow$43.5\%)} \\
    \midrule
    Hydra-MDP~\cite{li2024hydra} & 40.9 \\
    + Smoothing (BM)              & 38.8 ($\downarrow$\phantom{0}5.1\%) \\
    + Re-scoring w/ DrivoR        & 37.3 ($\downarrow$\phantom{0}8.8\%) \\
    + Re-scoring w/ GTRS-A          & 40.3 ($\downarrow$\phantom{0}1.5\%) \\
    + Re-scoring w/ SparseDriveV2 & \phantom{0}9.5 ($\downarrow$76.8\%) \\
    + Search w/ GTRS-A scorer       & 38.1 ($\downarrow$\phantom{0}6.8\%) \\
    + Search w/ SparseDriveV2 scorer & 29.6 ($\downarrow$27.6\%) \\
    \rowcolor{blue!10}
    \textbf{+ \ours{} (search w/ DrivoR scorer)} & \textcolor{blue}{49.7 ($\uparrow$21.5\%)} \\

    \midrule
    DrivoR~\cite{kirby2026drivor} & 54.6\\
    + Smoothing (BM)              & \rev{54.5 ($\downarrow$ 0.2\%)} \\
    + Re-scoring w/ GTRS-A        &  \rev{38.2 ($\downarrow$30.0\%)}  \\
    + Re-scoring w/ SparseDriveV2 &  \rev{39.0 ($\downarrow$28.6\%)}  \\
    + Search w/ GTRS-A scorer       & 24.7 \rev{($\downarrow$54.8\%)} \\
    + Search w/ SparseDriveV2 scorer & 40.1 \rev{($\downarrow$26.6\%)} \\
    + Safeguards w/ GTRS-A scorer        & 54.2 \rev{($\downarrow$\phantom{0}0.7\%)} \\
    + Safeguards w/ SparseDriveV2 scorer   & 54.8 \rev{($\uparrow$\phantom{0}0.4\%)} \\
    \rowcolor{blue!10}
    \textbf{+ \ours{} (search w/ DrivoR scorer)} & \textcolor{blue}{56.3 ($\uparrow$\phantom{0}3.1\%)} \\
    \bottomrule
  \end{tabular}
  }

\end{table}

\subsection{The gains require search and a generalizing scorer}
\label{sec:source}
Two factors explain \ours's gains: they come from search discovering new trajectories rather than re-ranking or smoothing existing ones, and they only appear with a scorer that remains accurate off the proposal distribution. We verify both in \autoref{tab:w_wo_MPC_small}, with \rev{iPad \cite{guo2025ipad}, Hydra-MDP \cite{li2024hydra}, and DrivoR~\cite{kirby2026drivor} as base planners} on \texttt{navhard-two-stage}.

\noindent\textbf{Search versus re-ranking and smoothing.}
We compare \ours against two simpler modifications of the base planner. \emph{Smoothing (BM)} maps the selected trajectory through the bicycle model to remove jitter without changing its route; it brings little improvement (iPad $+0.9\%$, Hydra-MDP $-5.1\%$, DrivoR $-0.2\%$), showing the gains are from better trajectories, {as seen in~\autoref{fig:success_failure_comparison}}. \emph{Re-scoring} re-ranks the original proposals with a scorer, without new trajectories. When using DrivoR's scorer, re-ranking is inconsistent: it improves iPad ($+31.4\%$) but hurts Hydra-MDP ($-8.8\%$). 
In contrast, \ours consistently improves performance ($+43.5\%$ and $+21.5\%$). \rev{\ours thus discovers} better trajectories outside the proposal set, \rev{rather than merely reordering or smoothing existing ones}.

\noindent\textbf{The scorer must generalize.}
We also run the search with different scorers as \rev{the reward}. \rev{Hydra-MDP shares its vocabulary with GTRS-A: the GTRS-A scorer roughly preserves performance when re-scoring it (40.3 vs.\ 40.9) but fails once the search explores new trajectories (38.1).} Only the DrivoR scorer \cite{kirby2026drivor} improves performance; GTRS-A \cite{li2025gtrs} and SparseDriveV2 \cite{sun2026sparsedrivev2} reduce it below baseline (e.g., iPad drops to 23.9 with GTRS-A). We attribute this to their training setup. GTRS-A and SparseDriveV2 are trained to score a fixed vocabulary of precomputed trajectories, so they receive no supervision away from that set and need not behave sensibly there. \rev{DrivoR's scorer}, by contrast, is trained jointly with a trajectory generator while remaining disentangled from it, forcing it to score freely decoded trajectories rather than a fixed grid. We hypothesize this is why it remains reliable on the off-proposal trajectories explored by our search, making it the only scorer \rev{among those tested} suitable as a reward. \rev{The same holds on DrivoR itself: searching with the GTRS-A or SparseDriveV2 scorer drops it from 54.6 to 24.7 and 40.1 EPDMS. {Weaker scorers can still be used} but with safeguards that make the search confidence-aware: the search stops early when the elite confidence falls below 0.8 of its iteration-0 value, and the refined trajectory is accepted only if the base planner's own scorer also rates it higher. These safeguards recover 24.7 $\to$ 54.2 (GTRS-A) and 40.1 $\to$ 54.8 (SparseDriveV2), back to the level of the base planner (54.6).}

\subsection{Computation Footprint and Choice of Optimizers}
\label{sec:footprint}

\definecolor{baseplanner}{HTML}{D3D3D3}
\definecolor{scorerbb}{HTML}{4D4D4D}
\definecolor{ttoblue}{HTML}{2E7DE8}

\begin{figure}[h]
\centering

\resizebox{\linewidth}{!}{
\begin{tikzpicture}
\begin{axis}[
    ybar stacked,
    bar width=22pt,              %
    x=1.35cm,                    %
    height=5.5cm,
    ymin=0, ymax=950,
    ytick={0,100,500},
    yticklabel style={font=\scriptsize},
    ylabel={ms / sample},
    ylabel style={
        at={(ticklabel cs:0.85)},
        anchor=south,
        yshift=-15pt,
        font=\scriptsize,
        inner sep=0pt
    },
    symbolic x coords={DrivoR,iPad,RAP,ZTRS,GTRS-A,Hydra-MDP},
    xtick=data,
    xticklabels={%
        {DrivoR\,\cite{kirby2026drivor}},
        {iPad\,\cite{guo2025ipad}},
        {RAP\,\cite{feng2025rap}},
        {ZTRS\,\cite{li2025ztrs}},
        {GTRS-A\,\cite{li2025gtrs}},
        {Hydra-MDP\,\cite{li2024hydra}}},
    xticklabel style={
        font=\scriptsize,
        rotate=30,
        anchor=north east,
        yshift=-1pt,
        xshift=-5pt,
    },
    ymajorgrids=true,
    grid style={dashed,gray!40},
    legend style={at={(0.56,0.95)},anchor=north west,draw=black,fill=white,font=\scriptsize},
    legend cell align=left,
    enlarge x limits=0.12,       %
    clip=false,
    tick label style={font=\scriptsize},
    label style={font=\scriptsize},
    every node near coord/.append style={anchor=center, font=\scriptsize},
]
\addplot+[fill=baseplanner,draw=black,
    point meta=explicit symbolic,
    nodes near coords, every node near coord/.append style={text=black, font=\scriptsize},
    nodes near coords align={anchor=center}] coordinates {
    (DrivoR,1.2) [] (iPad,9.2) [] (RAP,676.7) [676.7]
    (ZTRS,108.1) [108.1] (GTRS-A,91.8) [91.8] (Hydra-MDP,181.1) [181.1]};
\addlegendentry{Base planner}

\addplot+[fill=scorerbb,draw=black,
    point meta=explicit symbolic,
    nodes near coords, every node near coord/.append style={text=white, font=\scriptsize},
    nodes near coords align={anchor=center}] coordinates {
    (DrivoR,101.4) [101.4] (iPad,107.7) [107.7] (RAP,101.2) [101.2]
    (ZTRS,100.9) [100.9] (GTRS-A,101.2) [101.2] (Hydra-MDP,100.7) [100.7]};
\addlegendentry{Scorer backbone}

\addplot+[fill=ttoblue,draw=black] coordinates {
    (DrivoR,1.9) (iPad,1.9) (RAP,4.6) (ZTRS,19.4) (GTRS-A,20.4) (Hydra-MDP,19.1)};
\addlegendentry{\ours}

\node[below,font=\scriptsize,yshift=1.5pt] at (axis cs:DrivoR,0)    {104.5};
\node[below,font=\scriptsize,yshift=1.5pt] at (axis cs:iPad,0)      {118.8};
\node[below,font=\scriptsize,yshift=1.5pt] at (axis cs:RAP,0)       {782.5};
\node[below,font=\scriptsize,yshift=1.5pt] at (axis cs:ZTRS,0)      {228.4};
\node[below,font=\scriptsize,yshift=1.5pt] at (axis cs:GTRS-A,0)    {213.4};
\node[below,font=\scriptsize,yshift=1.5pt] at (axis cs:Hydra-MDP,0) {300.9};

\node[black,font=\scriptsize,anchor=west,xshift=9pt,yshift=3pt] at (axis cs:iPad,4.6) {9.2};

\node[above,ttoblue,font=\scriptsize,yshift=-2.5pt] at (axis cs:DrivoR,104.5)    {1.9};
\node[above,ttoblue,font=\scriptsize,yshift=-2.5pt] at (axis cs:iPad,118.8)      {1.9};
\node[above,ttoblue,font=\scriptsize,yshift=-2.5pt] at (axis cs:RAP,782.5)       {4.6};
\node[above,ttoblue,font=\scriptsize,yshift=-2.5pt] at (axis cs:ZTRS,228.4)      {19.4};
\node[above,ttoblue,font=\scriptsize,yshift=-2.5pt] at (axis cs:GTRS-A,213.4)    {20.4};
\node[above,ttoblue,font=\scriptsize,yshift=-2.5pt] at (axis cs:Hydra-MDP,300.9) {19.1};

\end{axis}
\end{tikzpicture}
}

\caption{\textbf{Per-sample inference cost across methods.} The bar length is the total inference time per sample, decomposed into the base planner (light gray), the scorer backbone (dark gray), and \ours (blue).}
\label{fig:cost}
\end{figure}

  \begin{figure*}[t]
      \centering
        \scriptsize
        \setlength{\tabcolsep}{2pt}
      \begin{subfigure}[T]{0.28\linewidth}
            \caption{\textbf{\ours component ablation.}}
            \resizebox{\linewidth}{!}{%
          \begin{tabular}{@{}l|crrrr|c@{}}
            \toprule
            \textbf{iPad} & \rot{warm-start} & \rot{bicycle} & \rot{scorer} & \rot{comfort} & \rot{anchor} & EPDMS$\uparrow$  \\
            \midrule
            \textit{Vanilla} &  &  & & &   & 34.7 \\
            \textit{normal init.}     &            & \checkmark & \checkmark & \checkmark  & \checkmark& 42.7 \\
            \textit{traj. space}      & \checkmark &            & \checkmark & \checkmark & \checkmark & 36.4  \\
            \textit{w/o scorer reward}       & \checkmark & \checkmark &            & \checkmark  & \checkmark& 35.7 \\
            \textit{w/o comfort term}     & \checkmark & \checkmark & \checkmark &            & \checkmark & 46.9 \\

            \textit{w/o anchor term}     & \checkmark & \checkmark & \checkmark &      \checkmark      &  &  44.4\\
           
            \midrule
            \rowcolor{blue!10}
            \textbf{\ours} & \checkmark & \checkmark & \checkmark & \checkmark  & \checkmark & \textcolor{blue}{48.5} \\
            \bottomrule
        \end{tabular}
        }
                \label{tab:cem_ablation}
      \end{subfigure}
      \hfill
      \begin{subfigure}[T]{0.23\linewidth}
          \centering
          \includegraphics[width=\linewidth]{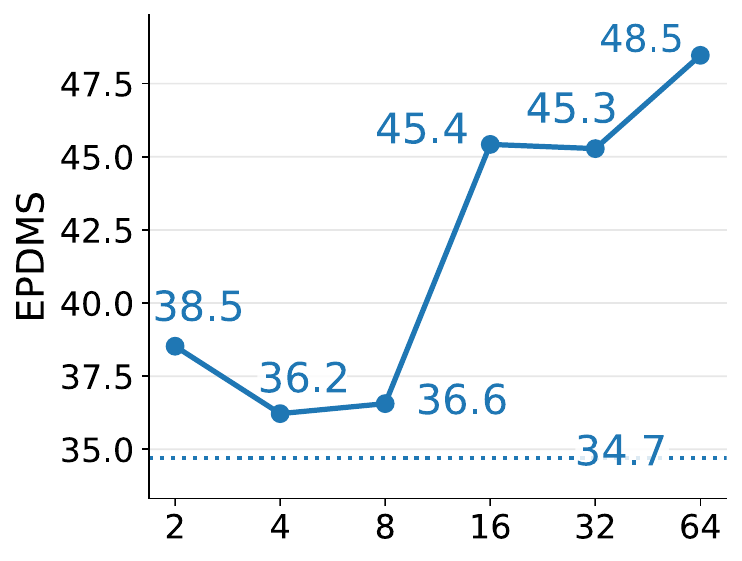}
          \caption{\textbf{Per-iteration candidates $M$}. $K=5$, $E=\max(1, M/8)$. }%
          \label{fig:mpc_ablation_proposals}
      \end{subfigure}
      \hfill
      \begin{subfigure}[T]{0.23\linewidth}
          \centering
          \includegraphics[width=\linewidth]{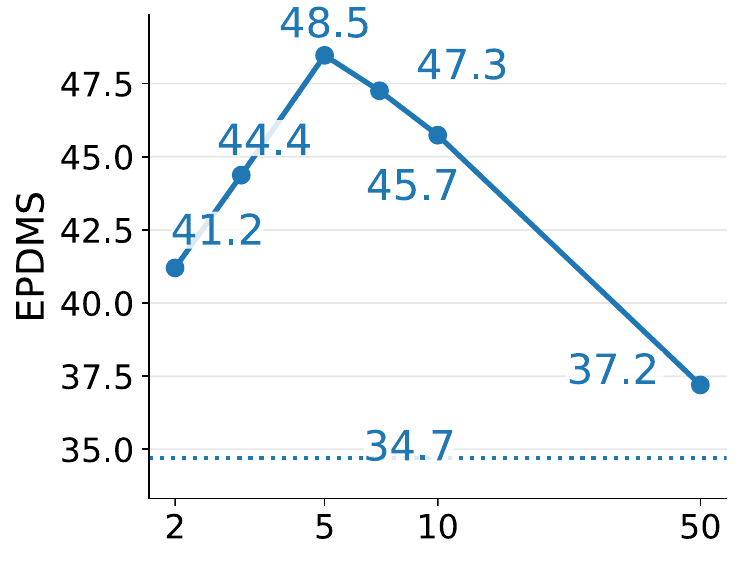}
          \caption{\textbf{CEM iterations $K$}. $M=64$, $E=8$.}
          \label{fig:mpc_ablation_iterations}
      \end{subfigure}
      \hfill
      \begin{subfigure}[T]{0.23\linewidth}
          \centering
          \includegraphics[width=\linewidth]{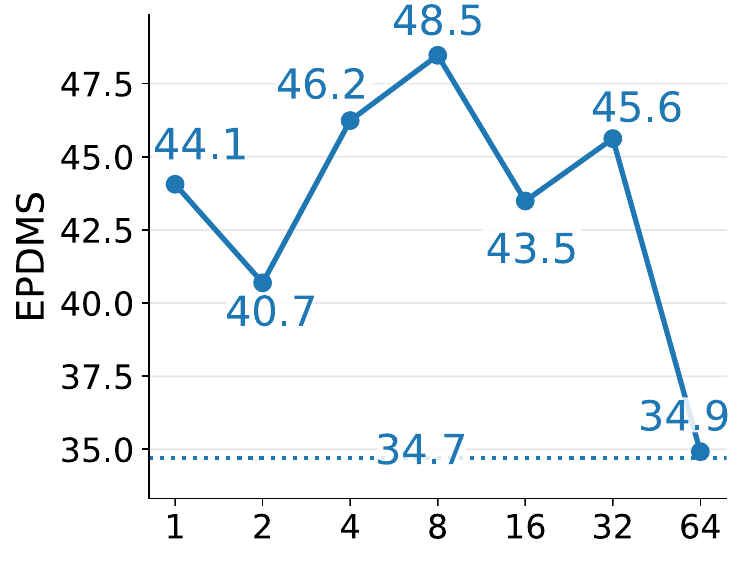}
          \caption{\textbf{Number of elites $E$}. $M=64$ and ${K=5}$. }
          \label{fig:mpc_ablation_elites}
      \end{subfigure}
      
        \caption{\textbf{Ablation study} of \ours with iPad on \texttt{warmup-two-stage}. }
      \label{fig:mpc_ablation}
  \end{figure*}

\autoref{fig:cost} reports the per-sample (\rev{batch size 1}) inference cost of each base planner + \ours, measured on a single NVIDIA A100. We split the execution time \rev{into} three parts: base planner, scorer backbone, and search itself. The scorer's perception backbone (a ViT-S encoder, around 100\,ms) is computed once per frame; the search then reuses these cached features and re-runs only the lightweight scoring head on each sampled trajectory. The search itself (blue) therefore adds only 1.9 ($K=5$) to 20.4\,ms ($K=100$), against total inference \rev{times of 105 to 783}\,ms.

The search cannot avoid running the scorer's perception backbone in the first place. In DrivoR the scorer is part of the base planner, sharing a single backbone: scoring reuses the features the planner already extracted, and the search is close to free (1.9\,ms on top of DrivoR's forward pass). When the scorer is external, it needs its own perception, redundant with the planner's, so features are extracted twice. This is the roughly 100\,ms scorer backbone carried by other planners in \autoref{fig:cost}. The lesson is to build well-generalizing scorers directly into the planner\rev{; strong test-time search then comes} essentially for free.

\subsection{Ablation Study}
\label{sec:ablation}

\noindent\textbf{\ours component ablation.} We ablate the different components in \ours with iPad~\cite{guo2025ipad}. $\bullet~$\textit{Warm-start.} We compare the warm-start initialization against a \textit{normal initialization}, with a mean set to zero and the standard deviation to one. $\bullet~$\textit{Bicycle model.} As an alternative \rev{to the} smoothed control space, \ours can also operate directly in the trajectory space. In this setting (\textit{traj.\ space}), we directly sample trajectories consisting of 8 future $(x,y)$ locations in bird's-eye view coordinates. $\bullet~$\textit{Objective terms.} We ablate the \textit{scorer}, \textit{comfort}, and \textit{anchor} terms by removing each in turn and comparing against both the \textit{vanilla} baseline and the \textit{full} configuration.

In \autoref{tab:cem_ablation}, we observe that the initialization from a standard normal distribution still achieves competitive performance (42.7 vs.\ 48.5), suggesting that \ours can generalize to a broader class of E2E planners that output only a single trajectory without associated scores.

Another important observation is that the scorer reward provides the primary optimization signal, leading to a substantial performance gain (48.5 vs.\ 35.7). The comfort and anchor terms provide smaller, complementary improvements (48.5 vs.\ 46.9, 44.4). Finally, optimization in the control space consistently outperforms optimization in trajectory space. We hypothesize that unlike \rev{the} naturally bounded control space (e.g., acceleration and yaw rate), trajectory coordinates are unbounded and may be more difficult to optimize effectively.

\begin{figure*}[t]
\centering
\includegraphics[width=0.19\textwidth, trim=20pt 0 0 0, clip]{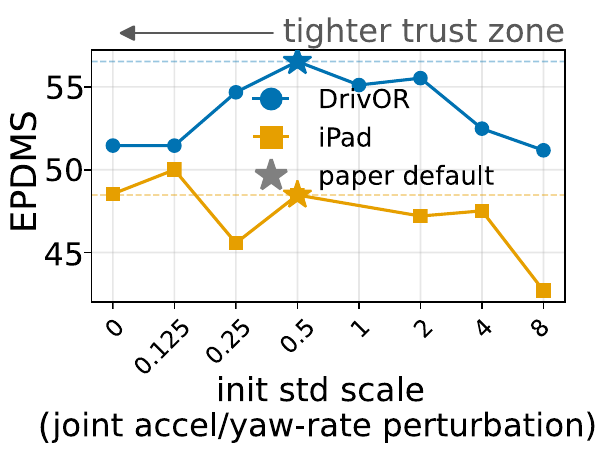}
\hfill
\includegraphics[width=0.19\textwidth, trim=20pt 0 0 0, clip]{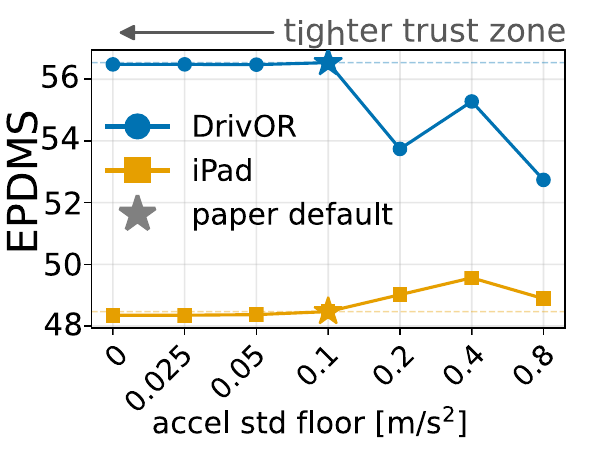}\hfill
\includegraphics[width=0.19\textwidth, trim=25pt 0 0 0, clip]{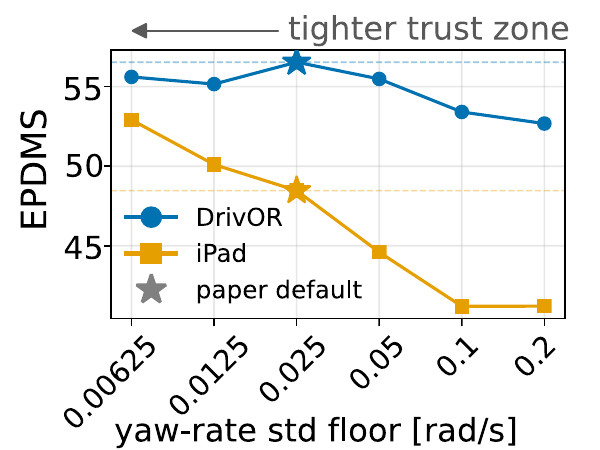}\hfill
\includegraphics[width=0.19\textwidth, trim=20pt 0 0 0, clip]{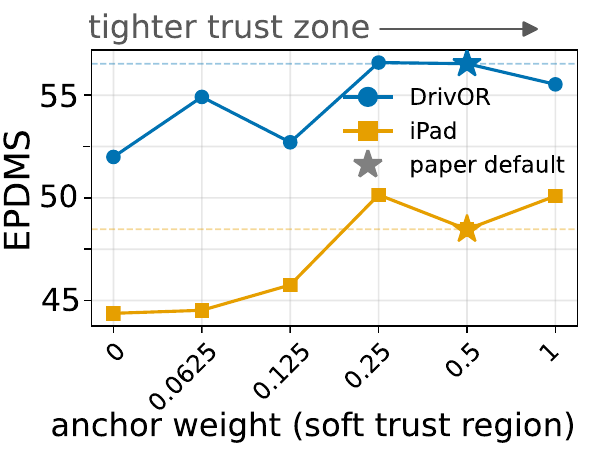}\hfill
\includegraphics[width=0.19\textwidth, trim=20pt 0 0 0, clip]{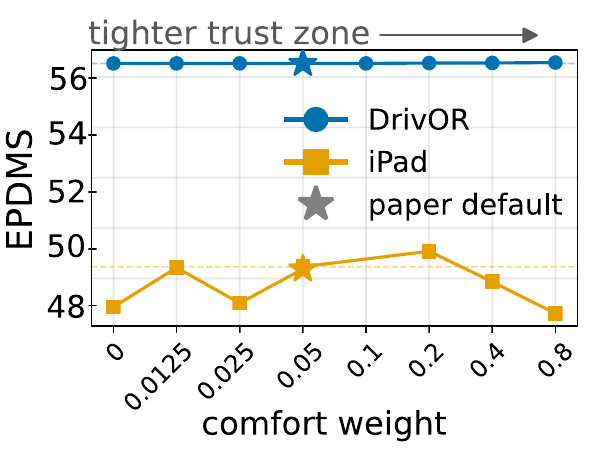}
\caption{\textbf{Hyper-parameter sensitivity.} EPDMS on \texttt{warmup-two-stage} when sweeping one CEM hyperparameter at a time, from left to right the initial-std scale $\beta$, the std floors $\epsilon_a$ and $\epsilon_\omega$, the anchor weight $\lambda_a$, and the comfort weight $\lambda_c$, for DrivoR and iPad with the DrivoR scorer; $\star$ and dashed lines mark the shared default.}
\label{fig:sensitivity}
\end{figure*}

\smallskip\noindent\textbf{Number of candidates $M$, iterations $K$, and elites $E$.}
Overall, \ours consistently outperforms the vanilla baseline across different hyperparameter settings, demonstrating its robustness and low sensitivity to hyperparameter choices. 
As shown in \autoref{fig:mpc_ablation_proposals}, performance generally improves as the number of sampled candidates increases. Interestingly, sampling only two control sequences already yields an improvement over the vanilla baseline. To maintain a low inference overhead, we cap the number of candidates at 64, resulting in \rev{about} 1.9\,ms runtime per execution.

In \autoref{fig:mpc_ablation_iterations}, we observe that the warm-start initialization enables rapid convergence, with iPad reaching peak performance after 5 iterations. The performance degradation observed with additional iterations highlights the imperfect correlation between the learned scorer and the true EPDMS objective. For instance, the scorer does not explicitly model extended comfort metrics that require temporal consistency across adjacent frames. Finally, \autoref{fig:mpc_ablation_elites} shows that selecting 8 elites empirically provides the best performance under a candidate budget of 64, corresponding to one-eighth of the sampled candidates. This result offers a practical guideline for choosing CEM hyperparameters.

\noindent\rev{\textbf{Sensitivity to trust-region hyperparameters $\beta$, $\epsilon_a$, $\epsilon_\omega$, $\lambda_a$, and $\lambda_c$.} The remaining hyperparameters define a trust region around the base proposal: the initial-std scale $\beta$, the std floors $\epsilon_a$ and $\epsilon_\omega$, the anchor weight $\lambda_a$, and the comfort weight $\lambda_c$. They are fixed once on \texttt{warmup-two-stage} and shared across planners. \autoref{fig:sensitivity} sweeps each one at a time on DrivoR and iPad with the DrivoR scorer. Trends are consistent across both planners, so no per-planner re-tuning is needed. Sharp drops appear only when exploration escapes the trust region: removing the anchor pull ($\lambda_a = 0$) costs 4.5 and 4.1 EPDMS for DrivoR and iPad, respectively, and a yaw-rate std floor $\geq 4\times$ the default costs 3.1 and 7.3.}

\begin{table}[t]
\centering
\caption{\textbf{Test-time optimization baselines.} On DrivoR~\cite{kirby2026drivor} (NAVSIM-v2, \texttt{navhard-two-stage}), gradient ascent (Adam, 10 steps) and MPPI (one softmax-weighted update) optimize the same controls as \ours{} (CEM) under the same frozen scorer, reward, and trust region. Latency and FLOPs are overheads over the base planner.}
\label{tab:tto_baselines}
\small
\setlength{\tabcolsep}{3pt}
\resizebox{\columnwidth}{!}{%
\begin{tabular}{lcccc}
\toprule
 & Base (DrivoR) & +Gradient (Adam) & +MPPI & +\ours{} (CEM) \\
\midrule
EPDMS        & 54.6  & +0.5  & +0.0           & \textbf{+1.7} \\
Latency (ms) & 102.6 & +99.7 & \textbf{+0.5}  & +1.9 \\
GFLOPs       & 351   & +0.98 & \textbf{+0.89} & +4.1 \\
\bottomrule
\end{tabular}%
}
\vspace{-5mm}
\end{table}

\noindent\rev{\textbf{CEM versus other optimizers.} \autoref{tab:tto_baselines} swaps CEM for two alternatives that optimize the same controls under the same frozen scorer, reward, and trust region on DrivoR: gradient ascent (Adam, 10 steps, learning rate 0.05) through the differentiable bicycle rollout and scorer, and MPPI~\cite{williams2017mppi} (temperature 0.3), a single softmax-weighted update in place of the iterated elite refits (CEM). CEM is the most effective (+1.7 EPDMS) for under 2\% latency and about 1\% FLOPs overhead. Gradient ascent stalls in local optima and serializes ten dependent forward-backward passes, while MPPI barely moves the sampling distribution. CEM also needs no differentiable scorer.}

\section{Conclusion}

\rev{End-to-end planners often score \textit{once} on a fixed candidate set. \ours, a general plugin that searches against the frozen scorer, improves seven base planners on NAVSIM benchmarks, and transfers zero-shot to closed-loop HUGSIM, at negligible cost and with no retraining. \ours also provides insights into scorer design: it succeeds only when the scorer stays accurate off the proposal distribution, which most current scorers do not. Two limitations point the way forward. \ours still warm-starts from a base planner, which a stronger scorer and optimizer could make unnecessary by searching from scratch. It also relies on a supervised disentangled scorer and its costly trajectory-level annotations, which self-supervised or world-model rewards could replace. Planners whose scorer is built to generalize would then get strong test-time search for free.}

{\small
\bibliographystyle{IEEEtran}
\bibliography{biblio}
}

\end{document}